\documentclass[a4, journal]{IEEEtranTIE}
\IEEEoverridecommandlockouts
\usepackage{cite}
\usepackage{amsmath,amssymb,commath,amsfonts}
\usepackage{graphicx}
\usepackage{textcomp}
\usepackage{url}
\usepackage{enumerate}
\usepackage{mathtools}
\usepackage{algorithm,algcompatible,algorithmicx}
\usepackage{xcolor}
\usepackage{adjustbox}

\usepackage[colorinlistoftodos]{todonotes}
\usepackage{caption,subcaption}
\usepackage{tabularx}
\usepackage{subcaption}
\usepackage{multirow} 
\usepackage[T1]{fontenc}
\usepackage[utf8]{inputenc}

\usepackage[colorlinks]{hyperref}

\DeclareMathOperator*{\argmin}{arg\,min}
\DeclareMathOperator{\atan2}{atan2}

\newcommand\red[1]{\textcolor{red}{#1}}

\def\BibTeX{{\rm B\kern-.05em{\sc i\kern-.025em b}\kern-.08em
		T\kern-.1667em\lower.7ex\hbox{E}\kern-.125emX}}
\DeclareUnicodeCharacter{2212}{-}
\UseRawInputEncoding

\begin{document}
	\title{Frontier-led Swarming: Robust Multi-Robot Coverage of Unknown Environments\\
		
		\thanks{The authors are with the School of Engineering and Information Technology, University of New South Wales, Canberra, Australia. The Commonwealthof Australia supported this research through a Defence Science Partnerships agreement with the Australian Defence Science and Technology Group.}
	}
	\author{\IEEEauthorblockN{Vu Phi Tran, Matthew A. Garratt, Kathryn Kasmarik, Sreenatha G. Anavatti}}
	\maketitle
	
	\begin{abstract}
		
	This paper proposes a novel swarm-based control algorithm for exploration and coverage of unknown environments, while maintaining a formation that permits short-range communication. The algorithm combines two elements: swarm rules for maintaining a close-knit formation and frontier search for driving exploration and coverage. Inspired by natural systems in which large numbers of simple agents (e.g., schooling fish, flocking birds, swarming insects) perform complicated collective behaviors for efficiency and safety, the first element uses three simple rules to maintain a swarm formation. The second element provides a means to select promising regions to explore (and cover) by minimising a cost function involving robots'  relative distance to frontier cells and the frontier's size. We tested the performance of our approach on heterogeneous and homogeneous groups of mobile robots in different environments. We measure both coverage performance and swarm formation statistics as indicators of the robots' ability to explore effectively while maintaining a formation conducive to short-range communication. Through a series of comparison experiments, we demonstrate that our proposed strategy has superior performance to recently presented map coverage methodologies and conventional swarming methods.

	\end{abstract}

\section{Introduction}
\red{\IEEEPARstart{A} common assumption in recent work related to multi-robot coverage path planning systems \cite{xu2014efficient,avellar2015multi,karapetyan2017efficient,nauta2020,huang2020}, is the availability of a map (that is, a known environment) at the planning stage. Furthermore, existing multi-robot coverage solutions for unknown environments, including graph-based approaches \cite{palacios2019equitable} and Voronoi-based approaches \cite{arslan2016voronoi}, perform coverage assuming individual robots are independent of each other \cite{avellar2015multi}. The failure of a single robot can cause incomplete area coverage. This paper relaxes the assumption that the structure of an environment is known in advance, and presents a cooperative swarming algorithm that incorporates search, coverage and self-organising formation control.} Motivated by emergent behaviors of animal aggregation found in insects, birds, fish, and other organisms, the interdisciplinary fields of swarm intelligence and robotics take advantage of complex yet efficient and safe self-organized swarm formations by building a system of apparently simple interaction rules that take local observable information and limited-range communication among proximate agents as inputs \cite{brambilla2013swarm}. The result is a system capable of robust and efficient coverage behaviour.  

The contributions of this paper are:

\begin{enumerate}
	\item A novel frontier-led swarming (FrontierSwarm) algorithm is introduced. This algorithm allows the robotic swarms to achieve fast area coverage while maintaining a close-knit formation to facilitate short-range communication.  
	\item A distributed architecture for implementing frontier-led swarming. In this architecture, each agent can directly exchange coverage and obstacle information with a set of neighbors without passing data through a central processing unit. Thus, the system is fully distributed and the loss of a subset of agents does not critically affect the operation of the remaining swarm members in completing the overall mission. 

	\item A comparative study benchmarking our frontier-led swarming algorithm against the frontier-based map coverage strategy without swarming and swarming algorithms without frontier search. The algorithms are tested on homogeneous and heterogeneous robot teams to cover a map as quickly as possible while maintaining a close, self-organized swarm formation to permit short-range communication. We introduce a combination of time, coverage and swarming metrics for the purpose of this comparative study. 
\end{enumerate}

The approach to swarming taken in this paper is based on Reynolds' boids model \cite{reynolds1987}. In contrast to other models such as the Vicsek model \cite{vicsek1995}, where flocking is generated by the simple combination of alignment interactions and random linear velocities, the Reynolds model \cite{reynolds1987} can be used to incorporate the collective motion behavior with other operations, such as reaching the goal and avoiding obstacles, by adding a set of external steering forces. Noticeably, although robots know their absolute locations at each step, boids methods do not require highly accurate positioning, which can be a drawback of several formation control strategies. The boids approach decomposes the complicated group motion behaviors into three simple steering rules (cohesion, alignment, or separation) at the individual agent level. Decision making is localised within neighbourhoods with only nearby agents within a relevant radius impacting the movement of any individual.

Moreover, as we will see in this paper, the boid swarming model offers a highly flexible framework that transfers between homogeneous and heterogeneous groups of robots. Recent swarming and flocking algorithms have tended to focus on homogeneous robot teams and initially assume perfect communication channels \cite{de2019bio,yu2019swarm,konda2020decentralized}. This paper presents a novel approach to exploration and coverage that extends to unknown environments where only limited, local communication is possible. We anticipate that this may be useful for environments where communication has been disrupted, for example following a disaster, or where communication channels are contested.

To achieve exploration and coverage, we adopt the concept of a \emph{frontier}, which is the border between unexplored and known areas \cite{gao2018improved,caley2019deep}. Frontier algorithms have previously been developed for search, and we adapt one in this paper for area coverage in an unknown environment.

The rest of this paper is organized as follows. In Section 2, a literature review of related work is presented, including our problem formulation and the preliminaries that form the underlying theory for our approach. Our frontier-led swarming algorithm and robotic architecture is introduced in Section 3. Our experimental environment, methodology and results are discussed in Section 4. Finally, concluding remarks are presented in Section 5. 

\section{Related Work}
In this section we first introduce the underlying mission we are addressing in Section A: area coverage in an unknown environment. We then review existing challenges related to coverage in Section B and existing approaches to search in an unknown environment in Section C. Finally we introduce the boid model that underlies our solution in Section D. 

\subsection{Area Coverage Problems}
Area coverage requires a group of agents or robots to visit every position in an environment. Typical real-world examples where area coverage must be applied include vacuum cleaning, painting and surveillance. To formally define the coverage problem a grid discretisation of a space is generally first defined with a certain resolution. Grids can be of different shapes including cells with triangular, square, hexagonal or diamonds shapes \cite{almadhoun2019}. Different definitions of coverage of the grid exist, including dynamic coverage and static coverage~\cite{kantaros2015distributed}. In \textbf{static coverage}, agents need to seek out optimal locations that form an appropriate geometry so that their network of sensors can cover the whole environment~\cite{zhong2011distributed}. However, in some practical missions, the environment cannot be fully covered by a static coverage geometric network due to the limitation in sensing range or too few sensors being available. In such cases, the problem of \textbf{dynamic coverage} is considered, where mobile robots with limited-range sensors continuously move and sample data, normally from unit spaces within a discretized domain until a predefined coverage level is attained~\cite{wang2010awareness}. In this paper, we focus on the dynamic coverage problem. We further assume that the environment to be covered is not known in advance. That is, the robots do not know the boundaries of the environment, nor the locations of any potential obstacles. 

In a simple formal definition of dynamic coverage, a 2D grid cell $j$ is covered by agent $A^i$ if $A^i$ was within cell $j$ at some point during the exploration \cite{masar2013biologically}. In that case, we say that $A^i$ has \textbf{directly covered} cell $j$. 

In a multi-agent system, if agent $A^i$ communicates the coverage of cell $j$ to a neighbour agent $A^k$, we say that agent $A^k$ has \textbf{indirectly covered} cell $j$. \textbf{Synchronised coverage} is achieved in a multi-agent system when each agent has directly or indirectly covered all grid cells of the map. In other words, each agent knows that each cell has been directly covered by some other agent.

Collaborative, complete exploration and coverage are two active research fields, where multi-robot systems are required to maintain situational awareness in unknown obstacle-cluttered scenarios. Solving these problems can benefit many potential applications, including environmental waste management~\cite{le2020multi}, patrolling~\cite{kuyucu2015} and search and rescue~\cite{albina2019}.

\subsection{Challenges Associated With Area Coverage and Existing Solutions}

In mobile robotics, the exploration task involves sensing the structure of unvisited areas and constructing a map. In contrast, the complete area coverage task aims at visiting or periodically re-visiting all necessary locations within a particular workspace~\cite{sharma2016survey}. Single-agent space coverage has been addressed by various means, such as genetic algorithms \cite{yakoubi2016}, coverage path planning  \cite{torres2016}, or reinforcement learning \cite{lakshmanan2020}. Algorithms using a single-agent may be optimal if sufficient time is available or if agents are expensive (e.g., planetary exploration missions). However, navigating and exploring a vast area using a single robot suffers from drawbacks, including long turnaround time, depletion of robot power resources, and poor quality and quantity of information collected \cite{almadhoun2019}. In particular, single-agent systems may not be an ideal choice for surveillance or disaster response when mission time is critical. 

In a modern multi-agent coverage system, information about the explored space is shared with nearby agents. Each agent maintains its own map and performs its own space coverage strategy. Many popular approaches rely on centralized control \cite{khaluf2019,schranz2020swarm} requiring high-bandwidth, reliable communication. Performance degrades when the number of agents increases or when communication bandwidth is limited or intermittent. Thus the multi-agent coverage task is widely considered in a distributed fashion \cite{Luo2017Neural,almadhoun2019,albina2019hybrid,song2020care}. Operating a distributed robotics framework considerably reduces the load of communication and enhances the resilience of system connectivity - important factors for large-scale networks \cite{li2008}. Another upside of the distributed system is that agents can have lower computational requirements and can consequently be of lower cost and more easily replaced~\cite{csahin2004swarm}.  

A body of existing work has considered coverage problems assuming a known environment. The environment may be known because the robot was given a map, or because it performed a pre-processing step to discover a map. The existence of a map permits planning based solutions to be pursued, in which way-points are identified offline. These way points are then ordered into a path plan before the robots engage in the actual activity of coverage. We consider these approaches here. 

\subsubsection{Viewpoint and Path Generation}

Viewpoint generation describes the production of way-points and poses of a vehicle during coverage. Viewpoint generation is an essential aspect of an area coverage problem because it provides the necessary guidance for the coverage path and the behaviour of the agents so that they can cover the structure or environment of interest as fast as possible.

\begin{figure}
	\centering
	\includegraphics[width=19.5pc]{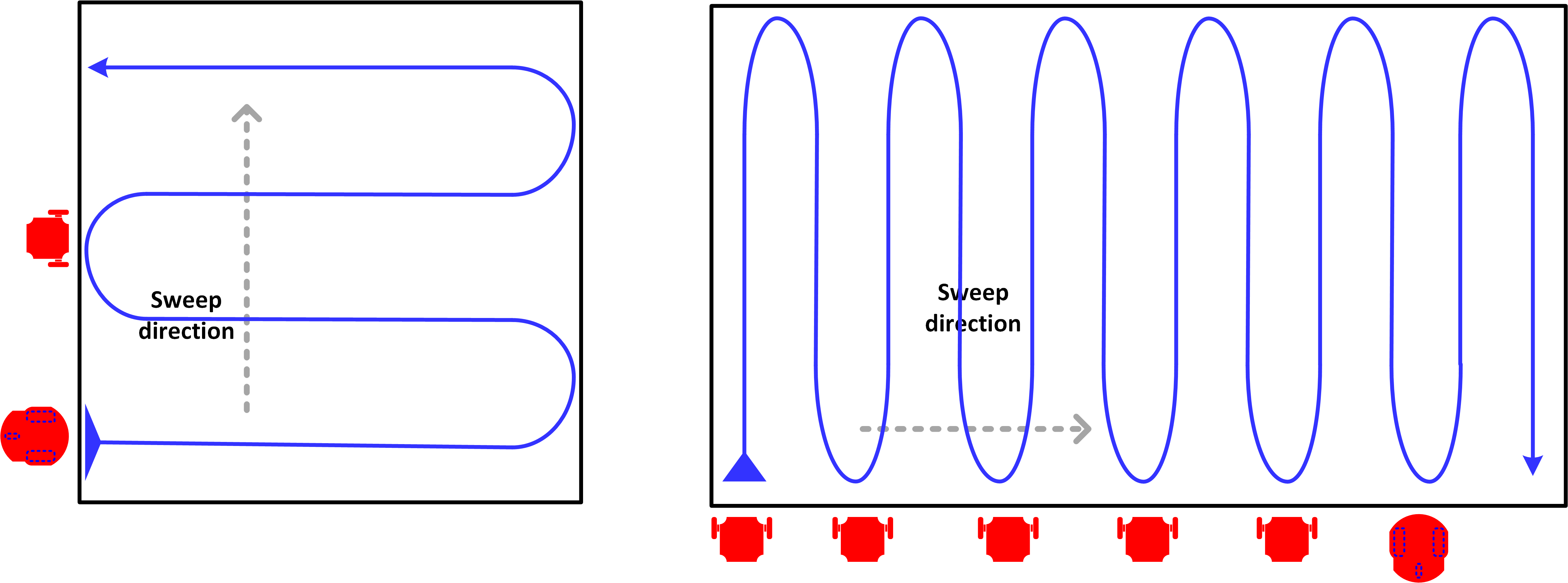}
	\caption{\footnotesize Routing-based coverage strategy. A rectangular field is covered using a back-and-forth motion along routes perpendicular to the sweep direction and turns along borders.}
	\label{fig:path_planning}
\end{figure}

Several previous studies propose \textbf{model-based} viewpoint generation approaches. The work in ~\cite{gautam2015cluster,modares2017ub} decomposed an area of interest into sub-regions and then assigns sub-regions to different robots. Many recent studies \cite{xu2014efficient,avellar2015multi,karapetyan2017efficient} propose a new solution for minimum-time coverage using a group of unmanned vehicles. The coverage problem is considered as a vehicle routing problem. In this method, the area of interest is decomposed into a set of sweeping line rows, as depicted in Fig. \ref{fig:path_planning}. The number of sweeps corresponds to the number of robots required to achieve coverage.

Although model-based techniques are quite useful, they are more suitable if one has a prior  model of the environment. In the cases of unobserved or large scale environments, whose characteristics are difficult to determine in advance, model-free viewpoints generation techniques are preferred.

Frontier-based methods are a range of state-of-the-art \textbf{model-free techniques} aimed at leveraging the benefits of tree-based frameworks and search algorithms. These approaches dominate recent literature on environment exploration and map construction tasks due to their accuracy and efficiency~\cite{almadhoun2016survey,gao2018improved,niroui2019deep}. The desirable unexplored sub-areas searched by a tree-based algorithm, so-called frontiers, are assigned as temporary destinations to visit by a single robot or group of robots. When there are no new frontiers left to detect, the entire environment is deemed to have been explored thoroughly ~\cite{sharma2016survey}. Some other techniques based on frontiers that reduce the computational requirements and exploration turnaround time can be found in \cite{caley2019deep,arvanitakis2016mobile,almasri2016trajectory}. Frontier-based exploration can perform well not only on a single mobile platform but also on multiple robots~\cite{cieslewski2017rapid,julia2012comparison}. 

Our study proposes a partially model-based method that leverages the frontier search algorithm's advantages to control the swarm coverage behaviors. Our approach includes a model-based component that only requires the environment's size as prior knowledge to form a spatial coverage-obstacle matrix within each individual robot's memory. The matrices' coverage status is shared locally to shape the swarm behaviors, leading to efficient exploration and coverage. However, we assume there is no prior knowledge of the obstacles located in the environment. Thus, a fully model-based method is inapplicable as it is impossible to compute a trajectory beforehand. To address this, a frontier-based (model-free) component is used in this paper to guide each robot to unexplored areas while obstacle avoidance behavior is achieved using a dynamic repulsive force technique.

\subsubsection{Search and Coverage}

 In existing work that combines aspects of the problem studied in this paper, Wang, Liang, and Guan~\cite{Wang2011Frontier} introduce a frontier search algorithm combined with particle swarm optimization (PSO) to improve the coverage capability of a swarm of mobile robots. The algorithm implements two working states: exploration and walking. In the exploration state, robots in a networked team discover their own non-overlapping sub-regions by travelling to local frontiers within such sub-areas. Once a robot explores all cells in its assigned area, it adapts a walking state to find another sub-area to explore using the PSO algorithm. The PSO algorithm aggregates three behaviours: a force representing the robot's direction in the previous time step, an attraction force toward the closest frontier cell in the map, and a dispersion force vector toward a cell with the minimum amount of other robots on its direction. When the robot reaches another sub-area, the exploration state starts again. This algorithm encourages swarm dispersion to cover more space within a period. In contrast, our approach maintains a close-knit formation to facilitate short-range communication among robots.

In a different approach, Cheng et al.~\cite{Cheng2009Distributed} designed a dynamic area coverage algorithm utilizing a swarm of autonomous robots, along with a leader-follower formation strategy. The robots possess only limited-range sensing and data processing capabilities. Due to these limitations, the authors present a hypothesis which states that using a flocking approach to create a formation of robots during the task would improve the exploration and coverage effectiveness and efficiency. The follower robots' location in the formation is dynamically adapted based on the movement of a leader robot in order to avoid obstacles. Meanwhile, a Braitenberg-motion-based coverage strategy is applied to cover the environment. Their findings demonstrate that the formation of multiple robots created by flocking behaviors is superior to an individual coverage strategy in terms of completion time and stability.

As described earlier in this section, for the area coverage problem, the cooperative robots need to discover unknown environments in minimum time and maintain proximity to support communication. The research in this paper is motivated by these goals.

\subsection{Reynolds' Boids Model of Swarming}

This section introduces the underlying swarming approach used in this paper: the boids model. A basic Boids swarming model \cite{reynolds1987} can be viewed as a type of rule-based reasoning to create swarming. The three fundamental rules are:
\begin{itemize}
\item Cohesion: An agent should move towards the average position of its neighbours.
\end{itemize}
\begin{itemize}
\item Alignment: An agent should steer to align itself with the average heading of its neighbours.
\item Separation: An agent should move to avoid collision with its neighbours.
\end{itemize}
The rules are generally implemented as forces that act on agents when a certain condition holds. Suppose we have a group of \(N\) agents \(A^1, A^2, A^3...A^N\). At time \(t\), agent \(A^i\) has a position, \(p_t^i\), and a  flocking velocity, \( v_{flock(t)}^i\). At each time step \(t\), the velocity of agent \(A^i\) is updated as follows:
\begin{equation}\label{eq:flock}
\begin{split}
v_{flock(t+1)}^i &= v_{flock(t)}^i + W_c v_{c(t)}^i + W_a  v_{a(t)}^i+ W_s v_{s(t)}^i \\ &+ (W_{av} v_{av(t)}^i) or (W_w v_{w(t)}^i)
\end{split}
\end{equation}
\(v_{c(t)}^i\) is the vector of cohesion force; \(v_{a(t)}^i\) is the vector of alignment force; \(v_{s(t)}^i\) is the vector of separation force; \(v_{w(t)}^i\) is the vector of the environment boundary (``wall") avoidance force; and \(v_{av(t)}^i\) is the vector of obstacle avoidance force. We distinguish the wall and obstacle avoidance forces in this paper because the environment boundary is known in advance, but the position of obstacles is not. Different positive weights \(W_c\), \(W_a\), \(W_s\), \(W_w\), and \(W_{av}\), are used to characterize the strengths of corresponding forces. Once a new velocity has been computed, the position of $p_{t+1}^i$ of agent $A^i$ in the next time step $t+1$ is computed as follows:
\begin{equation}
p_{(t+1)}^i=p_t^i+v_{flock(t+1)}^i                              \end{equation}

Formally, we can define a subset \(\mathcal{N}^i\) of agents within a certain range \(R\) of \(A^i\) as follows:
\begin{equation}
\mathcal{N}^i=\{A^k \;\; | \; k \neq i \; \land \; ||p_{t}^k-p_{t}^i||_2<R\}                         
\end{equation}
 
The number of agents in such a subset can be denoted by $|(\mathcal{N}^i)|$. Different ranges may be used to calculate cohesion, alignment and separation forces respectively, or other factors such as the communication range of a boid. The average position \(\bar{c}_t^i\) of agents in subset \((\mathcal{N}_c)_t^i\) whose positions are within the cohesion range \(R_c\) of \(A^i\) is calculated as:
\begin{equation}
\bar{c}_t^i=(\Sigma_k p_t^k )/|(\mathcal{N}_c)_t^i|                                                                   \end{equation}
The vector of cohesion force is the vector from the current position of agent $A^t$ towards this average position:
\begin{equation}
v_{c(t)}^i=\bar{c}_t^i-p_t^i                                     
\end{equation}
 
Similarly, we can calculate the average position \(\bar{s}_t^i\) of agents in subset \((\mathcal{N}_s)_t^i\) whose positions are within the separation range \(R_s\) of \(A^i\) as:
\begin{equation}
\bar{s}_t^i=(\Sigma_k p_t^k )/|(N_s)_t^i|                           
\end{equation}

The vector away from this position is calculated as:
\begin{equation}
v_{s(t)}^i=p_t^i - \bar{s}_t^i                                  \end{equation}
 
The average velocity \(v_{a(t)}^i\) of agents in subset \((\mathcal{N}_a)_t^i\) whose positions are within the alignment range \(R_a\) of \(A^i\), is calculated by:
\begin{equation}
v_{a(t)}^i=(\Sigma_k v_{flock(t)}^k )/|(N_a)_t^i|                    \end{equation}

These vectors can then be normalised and multiplied by their corresponding weights, before calculating the next velocity. The newly calculated velocity may be further normalised and then scaled by a speed value chosen in the range of $[V_{min}, V_{max}]$.


In this paper, a boundary wall collision avoidance velocity has been designed to keep the robots within a known rectangular area. The wall collision avoidance velocity $v_{w(t)}^i$ is defined by two components for a 2-D environment:
\begin{equation}
w^i_{x(t)}=\begin{cases}
-1~~~~if~x^i_t<\underline{x}\\
0~~~~~~if~\underline{x}\leq	x^i_t\leq \overline{x}\\
1~~~~~~if~x^i_t>\overline{x}
\end{cases}                                      
\end{equation}
\begin{equation}
w^i_{y(t)}=\begin{cases}
-1~~~~if~y^i_t<\underline{y}\\
0~~~~~~if~\underline{y}\leq	y^i_t\leq \overline{y}\\
1~~~~~~if~y^i_t>\overline{y}
\end{cases}                                      
\end{equation}
\begin{equation}
v_{w(t)}^i=(w^i_{x(t)},w^i_{y(t)}),
\end{equation}
where ($\underline{x}$, $\overline{x}$) and ($\underline{y}$, $\overline{y}$) denote the left, right, bottom, and top limits of the workspace.

Similar to the force calculation formula for environment boundaries, an obstacle avoidance force is calculated for obstacles, which is defined to be inversely proportional to the relative position between the i$^{th}$ robot and an unexpected obstacle $p^i_{ro}$. It can be computed as follows when the collision distance $R_{av}$ is violated:
\begin{equation}\label{eq:repulsive}
v^i_{av(t)}=\begin{cases}
-p^i_{ro(t)}~~~~if~p^i_{ro(t)} \leq R_{av}\\
0~~~~~~if~p^i_{ro(t)} > R_{av}\\
\end{cases}                                      
\end{equation}

In this paper we will demonstrate both simulated and real robots applying boid rules. Specifics of the sensors used to detect boundaries and unknown obstacles will be discussed later in the paper. We first present the formulation of our frontier-led swarming algorithm. 

\section{A Novel Frontier-led Swarming Algorithm}
This paper addresses the problem of coverage in a partially unknown environment with communication constraints. Specifically, the boundaries of the environment are pre-set, but the positions of obstacles within those boundaries are unknown. Communication is limited to short-range communication only. Here, we propose a distributed (decentralised) solution that requires a coverage matrix, search component and formation control. Therefore, this section is split into three parts: Section A introduces the space coverage matrix, Section B describes the frontier search algorithm, and Section C describes the swarm formation controller.

\subsection{Local Coverage and Obstacle Matrix}
Unlike a centralized setting where a common, global coverage matrix may be used by all agents, in this paper, every agent stores in its own memory, a 2D array $C_{A^i}$ composed of square grid cells of value $\varphi_{A^i}(X,Y)$ of appropriate resolution $\varepsilon$ covering the environment whose maximum and minimum coordinates $\overline{\cdot}$ and $\underline{\cdot}$ on the $x$ and $y$ axes are given in advance. We call this the local coverage-obstacle (CO) matrix of the agent. The map can be decomposed into dimensions as follows:
\begin{equation}\label{eq:dims}
dim(X,Y) = int((\overline{x} - \underline{x}, \overline{y} - \underline{y}) / \varepsilon)
\end{equation}
where $x$ and $y$ are positive integer numbers.
For each agent, all cells are first initialised as uncovered:
\begin{equation}
\varphi_{A^i}(X,Y) = 0, \;\; \forall X, Y \in \mathbb{N}.
\end{equation}

If the agent passes through an undiscovered or covered square cell ($X$, $Y$), then set:
\begin{equation}
\varphi_{A^i}(X,Y) = 1.
\end{equation}

Otherwise, if the agent senses an obstacle cell ($X$, $Y$) within a certain radius, then set:
\begin{equation}
\varphi_{A^i}(X,Y) = 2.
\end{equation}

When the distance between two agents $A^i$ and $A^k$ is smaller than a communication radius $R_{c}$, they exchange their local CO matrices. To do this, agent $A^i$ will update its internal CO matrix incorporating data from agent $A^k$'s CO matrix as follows:
\begin{equation}
\varphi_{A^i}(X,Y) (t+1) =  max(\varphi_{A^i}(X,Y)(t), \varphi_{A^k}(X,Y)(t)).
\end{equation}

This localised sharing of CO matrices aims to reduce network load and reduce the impact of issues such as loss of line-of-sight between pairs of robots. Short-range agent communications are well facilitated by swarming movements since the robots are continually pulled together due to the cohesion force that controls their movement (See Section 2.C). We now explain how a frontier search force can be incorporated with this swarming movement. 

\subsection{Frontier-led swarming}

\begin{figure}
	\centering
	\includegraphics[width=19.5pc]{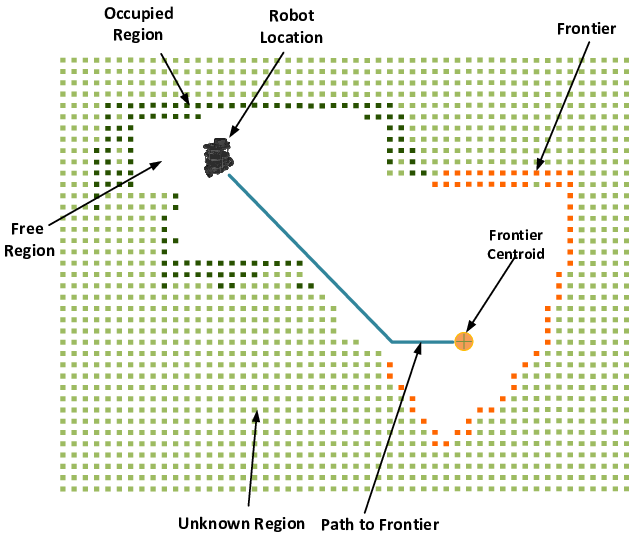}
	\caption{\footnotesize An example of a frontier, including visual representations of unknown, free and occupied regions.}
	\label{fig:fig1}
\end{figure}

In frontier search, the environment is approximated into a 2D or 3D grid map, where each cell possesses an occupancy value of 1 for explored, or 0 for unexplored. Frontiers are regions on the boundary that separates explored cells from unexplored cells. Any cell between an explored and unexplored cell is called a frontier cell. Groups of connected frontier cells are called frontier regions. Robots attempt to navigate to the nearest frontier region. The descriptions expressed above are detailed in Fig. \ref{fig:fig1}. 

Let $n^j_t$ denote the frontier cell $j$ at time step $t$. First, all valid frontier cells at time $t$ are collected into a list of frontier cells $\Gamma_t = [n^{0}_t,...,n^{J}_t]$, arranged in a queue data structure as described by Algorithm~\ref{al:frontier-cell}. A breadth-first search algorithm, illustrated in Algorithm~\ref{al:frontier-region}, is then implemented on each frontier cell in $\Gamma_t$ to map the frontier cells and relevant unexplored cells into a corresponding frontier region and add it in a set $\Xi_t = \{\zeta_t^1,...,\zeta_t^H\}, \;\; H \leq J$. Finally, the frontier regions are ranked to minimise the Euclidean distance $D_t^{j\rightarrow \zeta_t^h}$ from the robot's current position to the closest frontier cell of that region and the grid size $S_t^{j\rightarrow \zeta_t^h}$ of the frontier area that frontier cell represents. The utility function is defined as:
\begin{equation}\label{eq:3.1}
F_{\zeta_t^h} = \Psi_{D}D_t^{j\rightarrow \zeta_t^h} - \Psi_{S}S_t^{j\rightarrow \zeta_t^h},
\end{equation}
\begin{equation}\label{eq:3.2}
\tilde{\zeta}_t = \argmin_{\zeta_t^h \in \Xi_t} F_{\zeta_t^h},
\end{equation}
where $\Psi_{D}$ and $\Psi_{S}$ represent weighting parameters associated with the two terms. It can be seen that Eq. \ref{eq:3.1} is minimized when the distance is small and the frontier size is large. 

After an accessible frontier $\tilde{\zeta}_t$ is determined as a desired destination by (\ref{eq:3.2}), the algorithm then steers the robot towards that region's centroid. The search process is repeated by each robot until no more frontiers in the internal maps are found.  

\begin{algorithm}
\begin{algorithmic}
\renewcommand{\algorithmicrequire}{\textbf{Input:}}
\renewcommand{\algorithmicensure}{\textbf{Output:}}
\caption{Frontier cell search}\label{al:frontier-cell}
\REQUIRE current cell $c_t(X,Y)$ 
\ENSURE list of frontiers $\Xi_t$\\

\STATE Initialize a queue with $c_t(X,Y)$ as element $\Omega \gets c_t(X,Y)$
\STATE Initialize a list of frontiers $\Xi_t \gets \varnothing$ 
\WHILE{$\Omega$ is \textbf{not} empty}
    \STATE Dequeue a cell $c \in \Omega$
    \STATE Get set $\Pi$ including 4-connected cells of $c$
    \FOR{$c' \in \Pi$}
        \IF{$c'$ is a free cell}
            \STATE Add to queue $\Omega \gets c'$
        \ENDIF
        \IF{$c'$ is an unexplored frontier cell \textbf{and} $c'$ does not belong to any frontier}
            \STATE Find frontier region $\zeta_t(c')$ using Algorithm~\ref{al:frontier-region}
            \STATE Add to frontier list $\Xi_t \gets \zeta_t(c')$
        \ENDIF
    \ENDFOR
\ENDWHILE
\STATE \textbf{Return} $\Xi_t$
\end{algorithmic}
\end{algorithm}

\begin{algorithm}
\begin{algorithmic}
\renewcommand{\algorithmicrequire}{\textbf{Input:}}
\renewcommand{\algorithmicensure}{\textbf{Output:}}
\caption{Frontier region connectivity search}\label{al:frontier-region}
\REQUIRE frontier cell $n_t^j$
\ENSURE a frontier region contains frontier and unexplored cells as 8-connected region
 $\zeta_t(n_t^j)$\\

\STATE Initialize a queue with $n_t^j$ as element $\Phi \gets n_t^j$
\STATE Initialize frontier region $\zeta_t(n_t^j) \gets n_t^j$ 
\WHILE{$\Phi$ is \textbf{not} empty}
    \STATE Pop a cell $c \in \Phi$
    \STATE Get set $\Pi$ including 8-connected cells to $c$
    \FOR{$c' \in \Pi$}
        \IF{$c'$ is an unexplored free cell}
            \STATE Add to queue $\Omega \gets c'$
            \STATE Add to frontier region $\zeta_t(n_t^j) \gets c'$
        \ENDIF
    \ENDFOR
\ENDWHILE
\STATE \textbf{Return} $\zeta_t(n_t^j)$
\end{algorithmic}
\end{algorithm}

After the best frontier cell is selected, the attractive velocity toward the centre of the chosen frontier $p_{\tilde{\zeta}}$ acting on the relevant robot is obtained by:
\begin{equation}\label{eq:frontier}
v_{frontier} = \begin{cases}
p_{\tilde{\zeta}}-p^{r},~~if \; v_{s}^r = 0 \\
0,~~if \; v_{s}^r \neq 0\\
\end{cases},
\end{equation}

In case the same best cell is simultaneously selected by  two robots, a potential robot-robot collision can occur. To solve this issue, when two robots are closer than the critical radius $R_{s}$, they compute only the separation vector and neglect the frontier vector, as indicated in (\ref{eq:frontier}).  

\begin{figure}
	\centering
	\includegraphics[width=11.5pc]{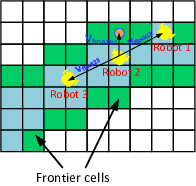}
	\caption{\footnotesize Fused velocity of the robot 2. Blue cells indicate the covered cells.}
	\label{fig:fused_force}
\end{figure}

The fused velocity vector, as shown in Fig. \ref{fig:fused_force}, is the summation of the frontier velocity and the flocking velocities:
\begin{equation}\label{eq:fused_vel}
v_{r} = \sum_{i=1}^{\mathcal{N}} v_{flock}^i +  v_{frontier}.
\end{equation}

Based on the fused velocity vector, the robot's linear and angular velocity can be computed as follows:
\begin{equation}\label{eq:control}
\begin{split}
V &= \sqrt{v_{rx}^2 + v_{ry}^2} \\
\omega &= k(\atan2(v_{ry}, v_{rx})-\theta),
\end{split}
\end{equation}
where ($v_{rx}, v_{ry}$) indicate the robot velocities on the $x$ and $y$ axes. The robot's angular velocity is $\omega$. Furthermore, $\theta$ is the robot's heading angle and $k$ is a constant gain.

\begin{figure}
	\centering
	\includegraphics[width=21.5pc]{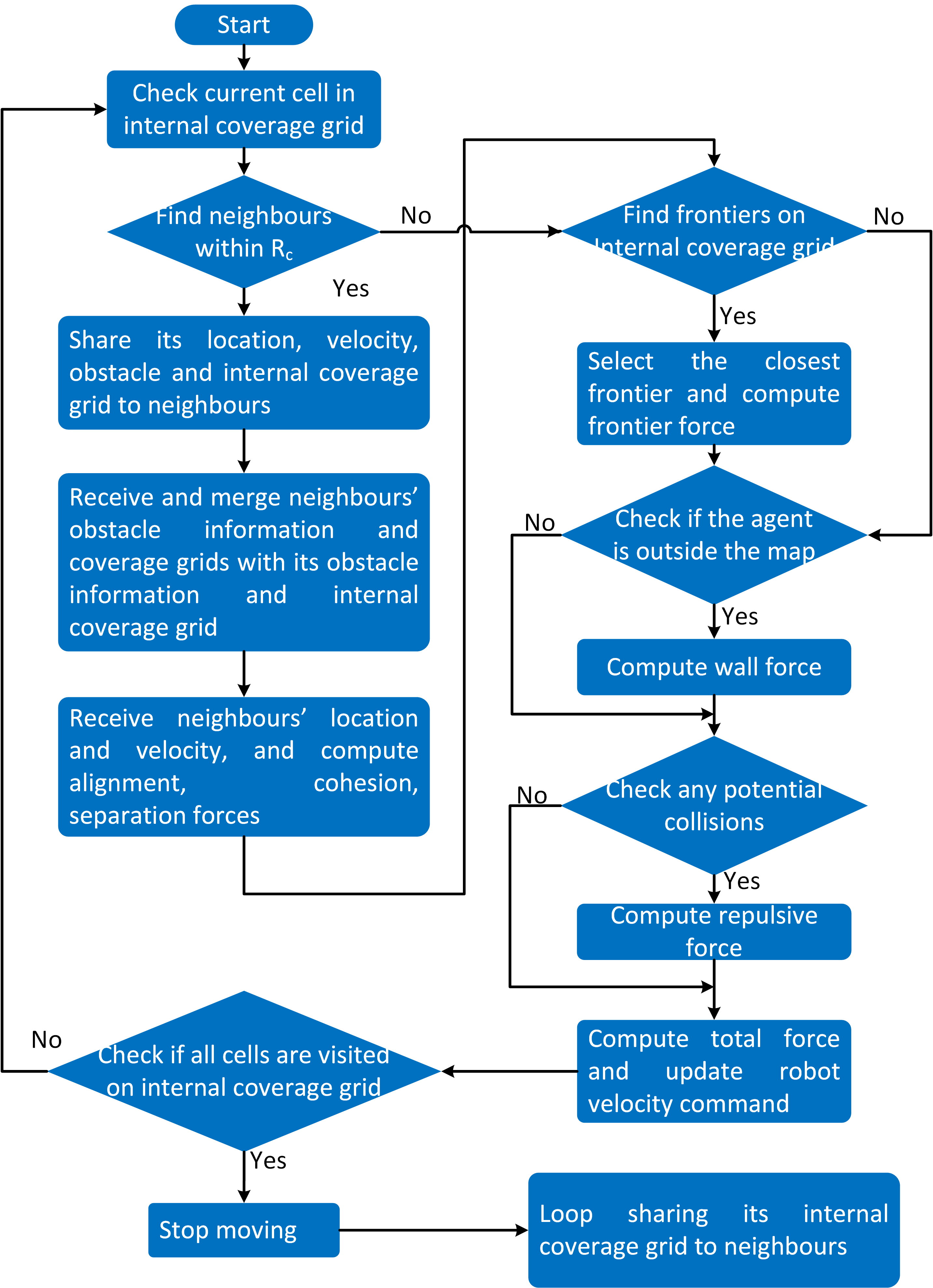}
	\caption{\footnotesize Flow chart of the proposed frontier-led swarming algorithm.}
	\label{fig:flowchart}
\end{figure}

Fig. \ref{fig:flowchart} illustrates a flowchart of the whole proposed control algorithm.

\subsection{Obstacle Avoidance}
In an uncertain or unknown environment, the robot can collide with another robot or an unexpected obstacle when traveling to the target point without a proper obstacle avoidance strategy. In this paper, we assume LiDar sensors provide data to implement a suitable  obstacle avoidance algorithm. We use a dynamic distance-based repulsive scale $\delta$ in this work to continuously convert the cohesion and alignment velocity vectors into the collision avoidance force as in (\ref{eq:repul_scale})-(\ref{eq:coh_wei}). By doing this, the cohesion force and the separation force acting on the robot will not be cancelled out when the robot is exploring nearby obstacles or in narrow passages.

\begin{equation}\label{eq:repul_scale}
\delta_{(t)}^i = \frac{R_{av} - p_{ro(t)}^i}{R_c}
\end{equation}
\begin{equation}
W_{av}^{'i} = W_{av}^i \delta_{(t)}^i
\end{equation}
\begin{equation}
W_{a}^{'i} = W_{a}^i max((1 - 2\delta_{(t)}^i),0)
\end{equation}
\begin{equation}\label{eq:coh_wei}
W_{c}^{'i} = W_{c}^i max((1 - 2\delta_{(t)}^i),0)
\end{equation}

\section{Experiments}
In this section we first describe the performance metrics against which we evaluate our algorithm in Section A. We then describe the robot setup in Section B and the environment setup in Section C. Comparative algorithms are described in Section D. Results and discussion are presented in Section E.

\subsection{Performance Metrics}

Four evaluation metrics are used for the different algorithms presented in this article. The first two metrics are concerned with the exploration and coverage performance of the system. The second two metrics are concerned with the self-organising formation control of the system. The first metric we will examine is the coverage percentage ($CP$) (\ref{eq:cover_per}). This metric tracks the percentage of the known region that has been visited by any robot.

\begin{equation}\label{eq:cover_per}
CP = \frac{|\Theta_e|}{|\Theta|}100\%
\end{equation}

where $|\Theta_e|$ and $|\Theta|$ are the number of covered cells and the total number of cells to be covered in this problem respectively.

The second metric is the turnaround time ($TT$). This metric summarises the coverage performance in terms of the time taken for the robots to indirectly cover the known environment, including mapping of all unknown obstacles.

The third metric is the `group' metric $G$ (\ref{eq:group_metric}), which estimates how closely a set of robots is clustered together (a lower value indicates greater grouping). The grouping metric $G$ is defined by Equation \ref{eq:group_metric} where $N_{r}$ is the number of the robots, $\bar{p}_{r}$ is the average position of the $N_{r}$ robots at the given time, and $T_0$ is the time that the robots start moving.

\begin{equation}\label{eq:group_metric}
G =  \sum_{t=T_0}^{T_0+15dt} \frac{\frac{1}{N_{r}} \sum_{i=1}^{N_{r}} ||p_{r}^{i} - \bar{p}_{r}||_2}{N_G},
\end{equation}
where $dt$ depicts the sample time. $N_G$ = 15 represents the sample length of the group metrics.

The fourth metric is the `order' metric $O$ (\ref{eq:order_metric}), which computes how similarly the robots are  aligned (a lower value indicates greater alignment). The order metric $O$ is defined by Equation \ref{eq:order_metric} below where $\bar{v}_{r}$ is the average velocity of the $N_{r}$ robots at the given time.

\begin{equation}\label{eq:order_metric}
O = \sum_{t=T_0}^{T_0+15dt} \frac{\frac{1}{N_{r}} \sum_{i=1}^{N_{r}} ||v_{r}^{i} - \bar{v}_{r}||_2}{N_O},
\end{equation}
Here, $N_O$ = 15 represents the sample length of the order metrics.

Five runs were conducted for each experiment, and the evaluation metrics' \textbf{mean and standard deviation} are computed and shown in all result tables in Section 4.E. Video data was also collected. All video demonstrations showing spatial coverage of Robot 1 in every experiment can be viewed at the following address: \url{https://youtu.be/4mDpP_TMjFU}; \url{https://youtu.be/XqiMdQJvkjQ}; \url{https://youtu.be/2t-X7_5FeIY}; and \url{https://youtu.be/xOJvz0ee0sg}.

\subsection{Experimental Setup of Robots}
\begin{figure}
	\begin{center}
		\begin{tabular}{cc}	
			\includegraphics[width=10.0pc]{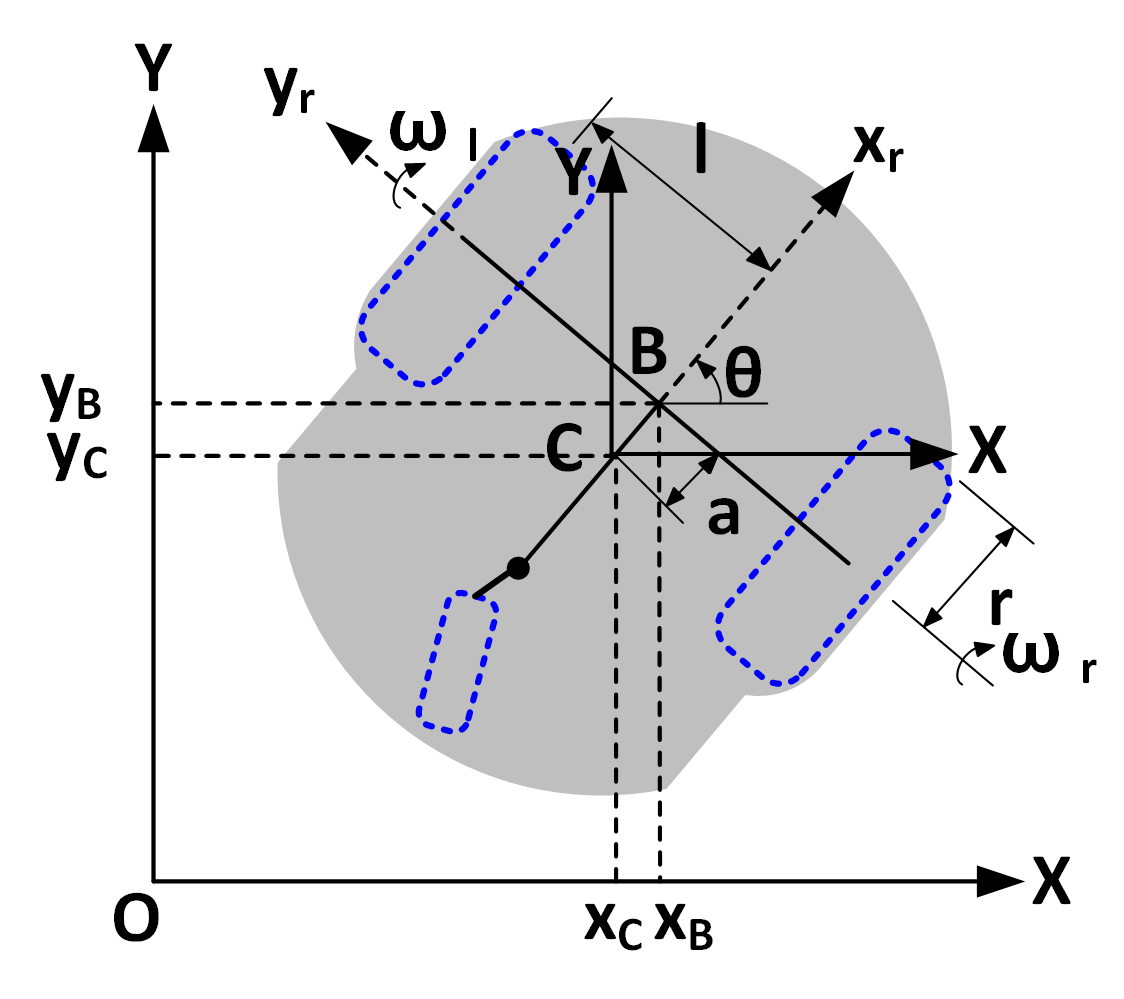} 
			\includegraphics[width=10.0pc]{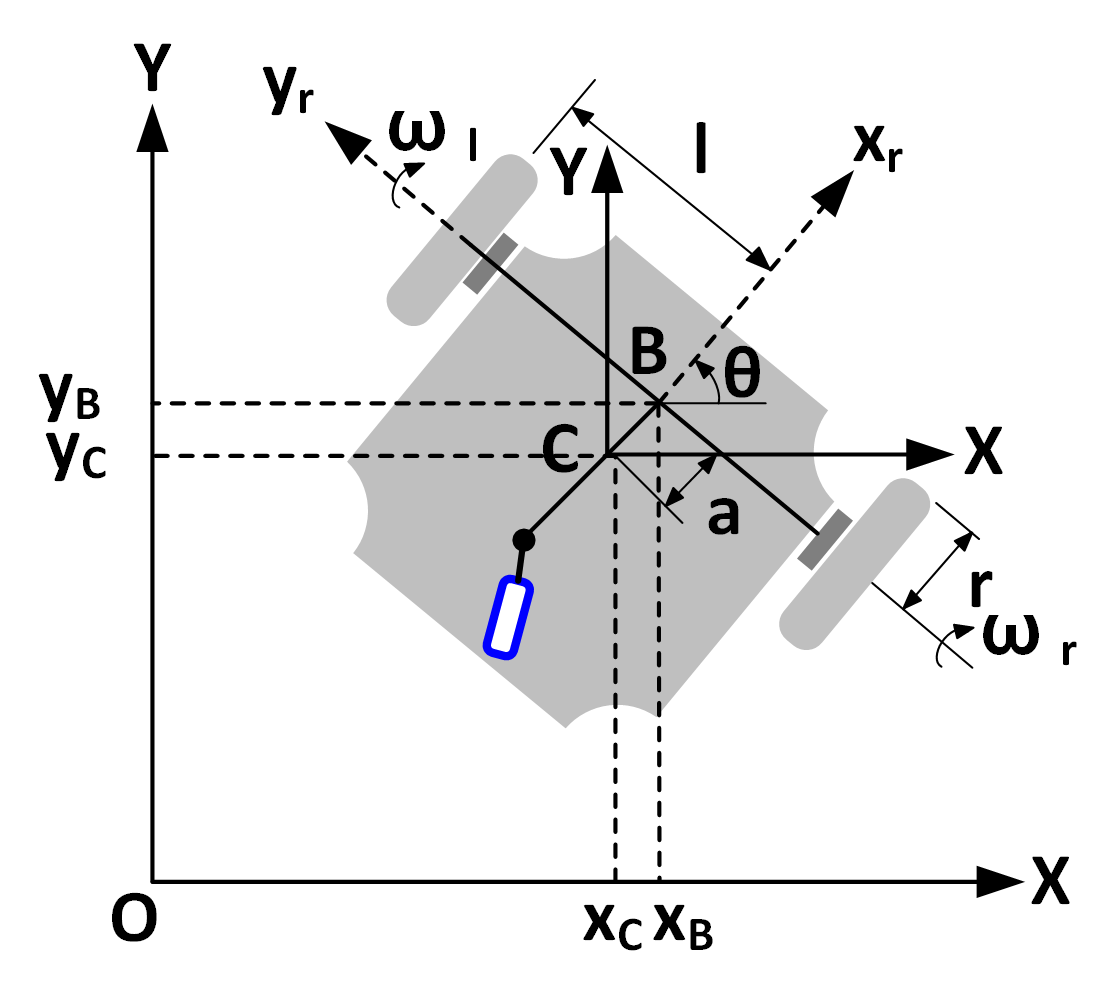} \\
			(a) \textit{Pioneer 3DX} ~~~~~~~~~~~~
			(b) \textit{TurtleBot Burger}\\
		\end{tabular}
		\caption{\footnotesize Schematic of two autonomous ground vehicles.}
		\label{fig:scheman}
	\end{center}
\end{figure}

The experiments in this paper use TurtleBot Burger and Pioneer 3DX robots. TurtleBots are small, reliable, low-cost two-wheel differential drive unmanned ground vehicle (UGV) platforms (see Fig. \ref{fig:scheman}). Pioneer 3DX is a three wheeled robot. In our experimental setup, the robots are controlled by on-board computers integrated with the Robot Operating System (ROS) and other compatible open-source software. We utilise 360-degree LiDAR sensors for obstacle detection.

To control these  UGVs, ROS supports two control input signals, namely yaw rate and linear velocity. These desired speeds are computed using Eq. \ref{eq:control}, and then used as the reference signals for inner loop PID controllers which control the servo motors driving the wheels of each robot type.

\begin{figure}
	\centering
	\includegraphics[width=21.5pc]{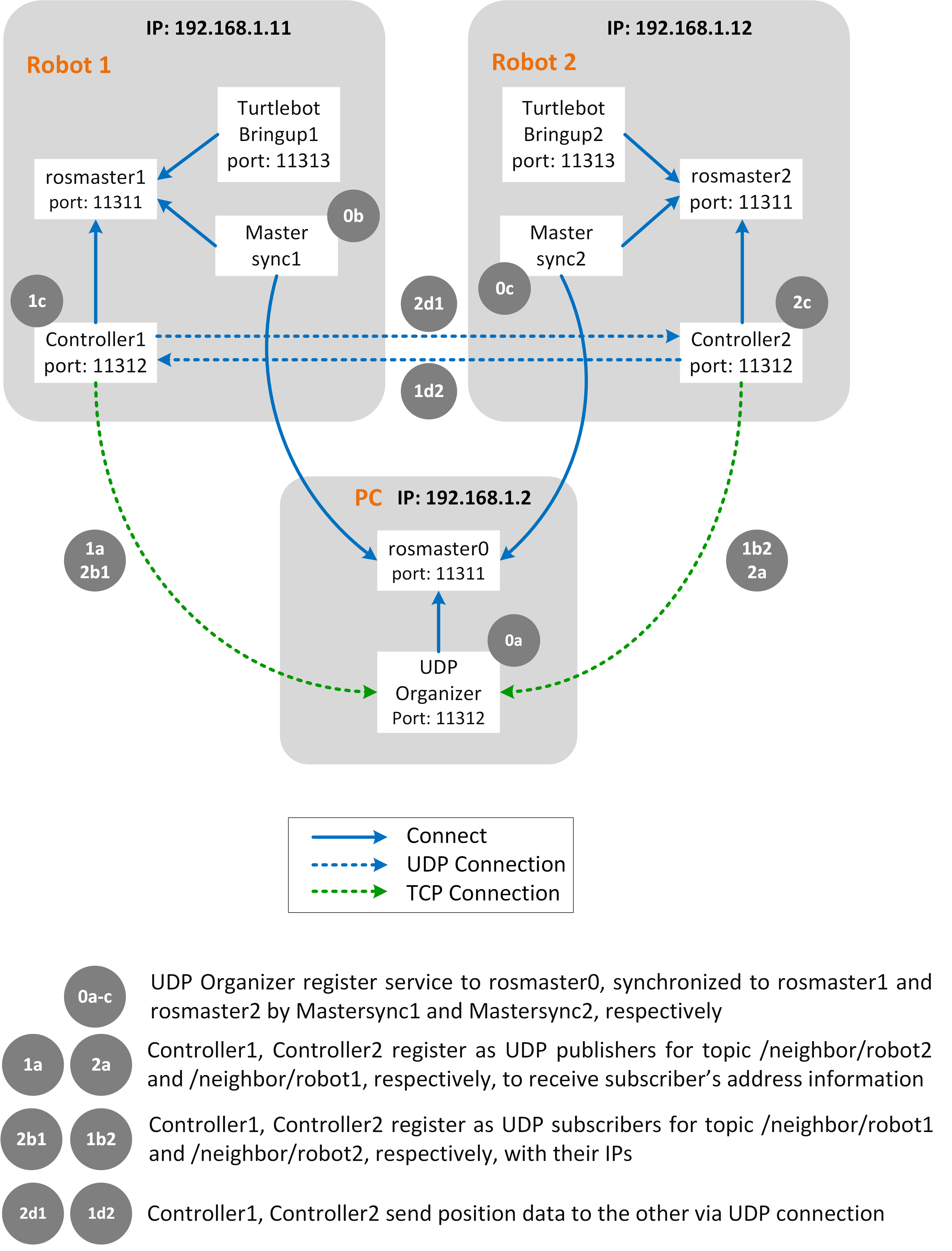}
	\caption{\footnotesize Distributed \textit{Multi-ROS Master} architecture for the swarming behavior of the cooperative ground vehicles. Only two robots are shown for simplicity.}
	\label{fig:multi_master_arch}
\end{figure}

In our real-time, controlled experiments, the behavior of all robots is captured by a VICON Tracker motion capture system and transmitted to each robot every 0.02s through the User Datagram Protocol (UDP) network protocol. In future real applications, we envisage this central localisation node would be replaced with a GPS or ultra-wideband (UWB) sensor mounted on each robot to provide the global position of each robot. The CO matrices for each robot are shared with neighbouring robots at an update rate of 0.01s. The sampling time is fixed at 0.02s. All proposed algorithms and the multi-ROS Masters were implemented on each robot (see Fig. \ref{fig:multi_master_arch}). Thus, if one of the robots fails due to a power outages or hardware problem, the remaining robots can still complete the coverage task. In our experimental simulations, the same settings are implemented using ROS and Gazebo tools.

Table \ref{tab:robot} shows the parameter set up for the physical experiments and the two different types of robots employed in this work. \red{These parameters were set and tuned manually using established rules of thumb for boid swarms \cite{reynolds1987}. Specifically, the cohesion force was set at a low value, with the separation and obstacle avoidance forces set stronger, but operating in a smaller radius. The separation radius was selected to maintain the physical safety of each robot, and other parameters were tuned manually. }

\begin{table}[h]
 \centering
\caption {TurtleBot Burger and Pioneer 3-DX Robot Parameters} 
\label{tab:robot}
  \centering
  \begin{tabular}{llll}
    \hline
     \textbf{Parameter} &  \textbf{Description} &  \textbf{Burger} &  \textbf{P3-DX}\\
      \hline
$W_{a}$ &  Alignment weight & 0.5 & 0.6\\
$W_{c}$ &  Cohesion weight & 0.23 & 0.21\\
$W_{s}$ &  Separation weight & 1.1 & 1.2\\
$W_{w}$ &  Boundary avoidance weight & 1.1 & 0.05 \\
$W_{f}$ &  Cohesion force weight & 0.08 & 0.03\\
$W_{av}$ &  Obstacle avoidance force weight & 1.1 & 1.1\\
$\Psi_{1}$ &  Frontier distance weight & 0.001 & 0.001\\
$\Psi_{2}$ &  Frontier size weight & 1.0 & 1.0\\
$R_{a}$ (m) &  Alignment radius & 1.5  & 1.5 \\
$R_{s}$ (m)&  Separation radius & 0.65 & 0.85 \\
$R_{c}$ (m)&  Cohesion radius & 1.5  & 1.5 \\
$R_{av}$ (m)&  Obstacle avoidance radius & 1.7  & 1.7 \\
$\overline{\dot{V}}$ (m.s$^{-1}$)&  Maximum linear speed & 0.25  & 0.25 \\
$\overline{\dot{\theta}}$ ($^{o}.s^{-1}$)&  Maximum angular speed & 0.9   & 0.9 \\
$\varepsilon$ &  CO matrix resolution & 0.3 m & 0.3 m\\
\hline
  \end{tabular}
\end{table}

\subsection{Experimental Environments}

Three experimental environments were established: (1) a small, empty square arena 3.6$\times$3.6m based on our VICON motion capture system laboratory; (2) a larger 4.4$\times$4.8m square arena with four boxes representing unknown obstacles in our VICON laboratory; and (3) a simulated, cluttered, urban-like environment. 

\begin{figure}
	\centering
	\includegraphics[width=18.5pc]{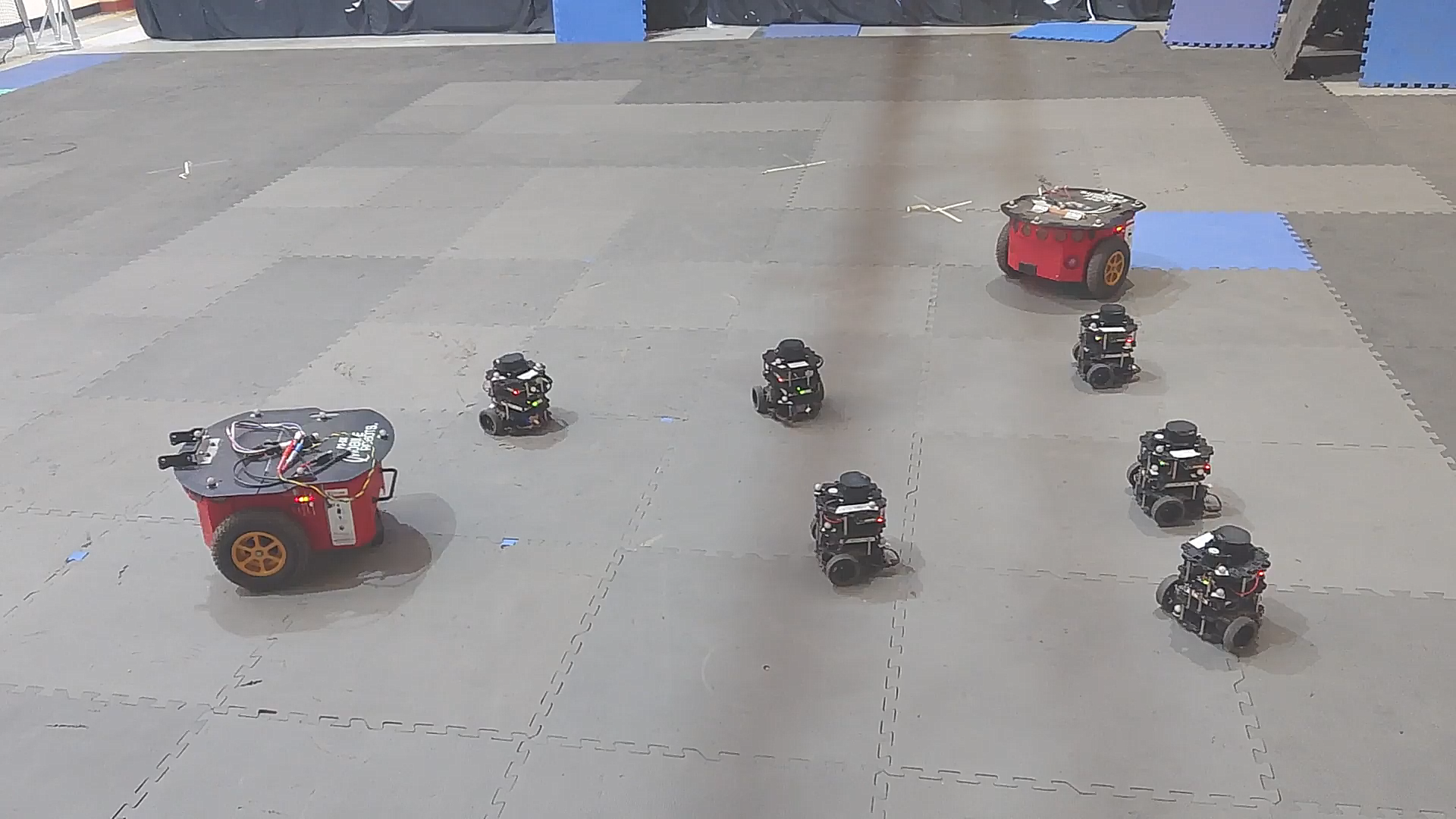}
	\caption{\footnotesize Small VICON enabled environment with a heterogeneous swarm of six Turtlebot and two Pioneer UGVs.}
	\label{fig:experim_area1}
\end{figure}

Environment 1 is shown in Fig. \ref{fig:experim_area1}. In this picture we can see a heterogeneous swarm of six TurtleBot and two Pioneer UGVs. This environment permits us to examine the performance of our algorithm on real robots in a known, controlled setting. 

\begin{figure}
	\centering
	\includegraphics[width=13.5pc]{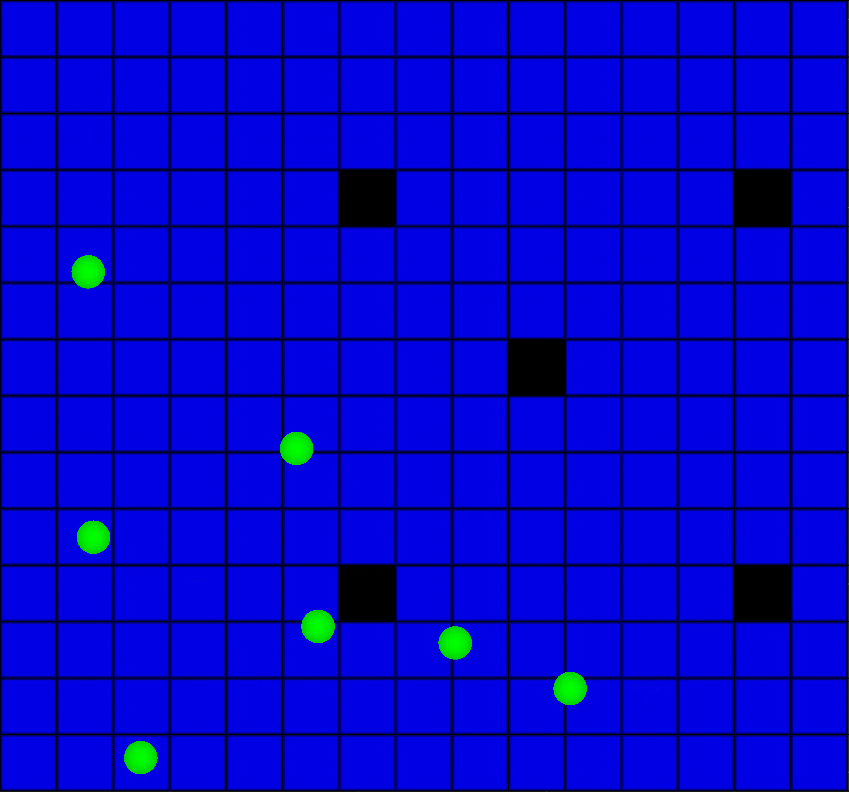}
	\caption{\footnotesize Large map region and obstacles. Green spheres describe the robot positions while blue squares depict the covered cells. Further, black squares are the obstacle cells.}
	\label{fig:ereal_map_ob}
\end{figure}

An abstract representation of Environment 2 is shown in Fig. \ref{fig:ereal_map_ob}. The static obstacles are represented by the black squares in Fig. \ref{fig:ereal_map_ob} and other subsequent figures. We will use this representation to visualise coverage behaviour in the results section of this paper. This environment permits us to examine the performance of our algorithm on real robots in an environment with unknown obstacles. 

\begin{figure}
	\centering
	\begin{subfigure}[b]{0.3\textwidth}
		\includegraphics[width=\textwidth]{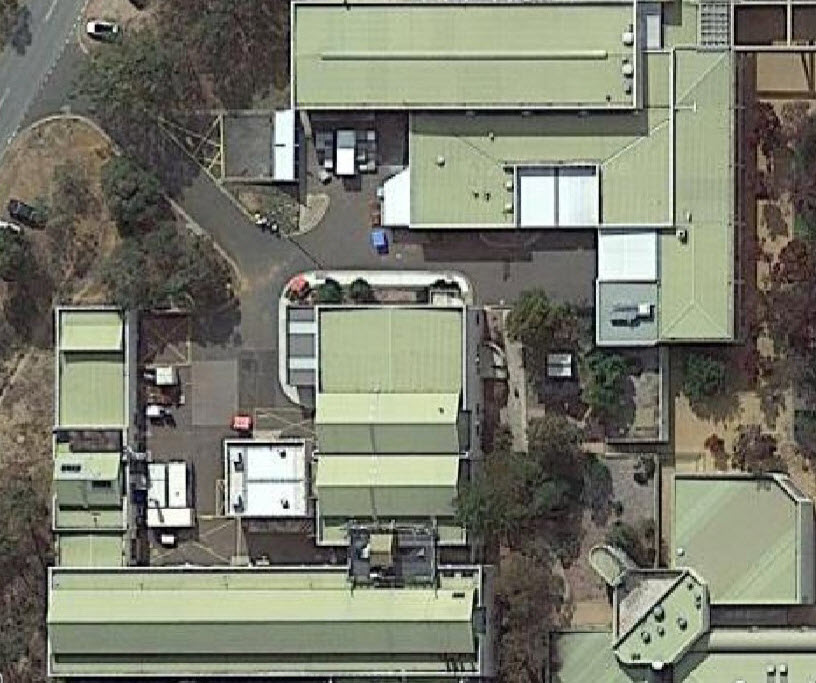}
		\caption{\textit{Real built environment.}}
		\label{fig:urban}
	\end{subfigure}
	\begin{subfigure}[b]{0.3\textwidth}
		\includegraphics[width=\textwidth]{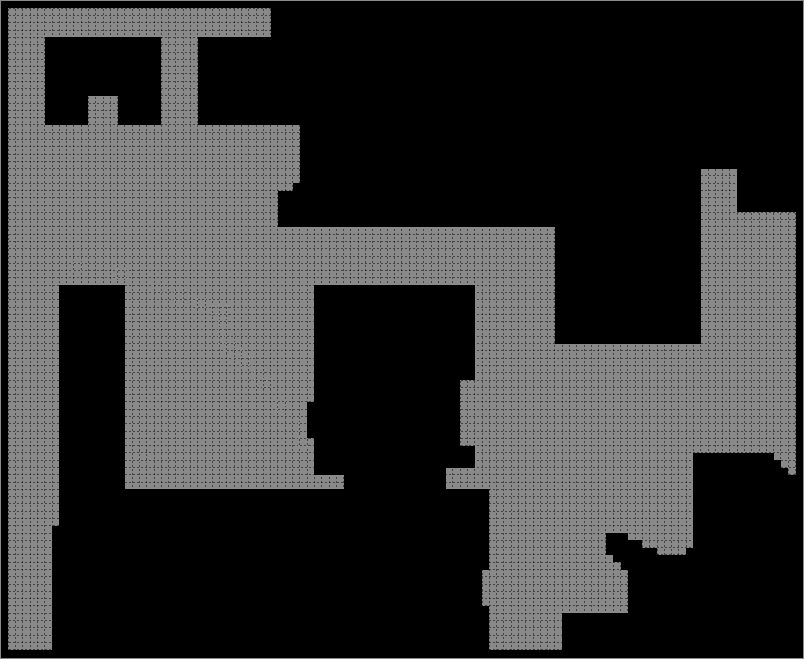}
		\caption{\textit{Virtual Gazebo environment.}}
		\label{fig:map_sim}
	\end{subfigure}
	\caption{\footnotesize Unknown complex urban-like environment.}\label{fig:urban_en}
\end{figure}

Environment 3 is a 2D large-scale urban-like scenario 120$\times$90m cluttered with obstacles simulated in a virtual Gazebo world (see Fig. \ref{fig:urban_en}). This environment enables us to examine the performance of our algorithm in a larger unknown environment. We use two scaled versions of this environment to investigate the impact of environment size on performance. 

For measurement purposes in this paper, we assume that once all frontier cells in the local map of any robot are covered directly or indirectly, a message `completed task' is sent to the remaining robots and the experiment ends. In future systems, such a message may also be sent to a base station or operator through a separate low-bandwidth longer range link. This message may also contain the complete coverage map, or the robot may be programmed to return to a set location to update other robots or the base station/operator with the final map. 

\subsection{Comparative Algorithms}

In this section, the performance of our Frontier-led Swarming (FrontierSwarm) algorithm is examined and compared with four alternatives. The first is a  Frontier (Frontier) algorithm without swarming. This approach uses the frontier force and separation rule only (to prevent robots colliding). The purpose of this comparative algorithm is to provide benchmarking against a state-of-the-art exploration method. 

The second comparative algorithm is a greedy (non-frontier) spatial coverage method with swarming (GreedySwarm). This approach uses the same boid rules and CO matrix. However, instead of using a frontier attraction force $v_{frontier}$ as described in Section 3, the final velocity is computed with an attractive force toward the nearest unexplored cell found by Algorithm \ref{al:strategy3} \cite{masar2013biologically}.  The purpose of this comparative algorithm is to confirm that frontier-led swarming provides an advantage over a more naive (greedy) exploratory approach. 

The third is swarming (Swarm) only without any frontier or other attractive component towards unexplored cells \cite{chibaya2008}. The purpose of this comparative algorithm is to provide a baseline for understanding the impact of an exploratory force (whether frontier-led or greedy) on swarming performance. 

Finally, a multi-vehicle routing-based area coverage approach \cite{avellar2015multi,karapetyan2017efficient} is also considered in the comparative experiments. This provides a way to compare performance to an approach that has zero coverage redundancy (but which requires a map of the environment). 

\begin{algorithm}
\begin{algorithmic}
\renewcommand{\algorithmicrequire}{\textbf{Input:}}
\renewcommand{\algorithmicensure}{\textbf{Output:}}
\caption{GreedySwarm cell selection using a greedy approach}\label{al:strategy3}
\REQUIRE current cell $c_{t}^0$; list of unexplored cells $\Theta_{u(t)}$; the map size; and the coverage matrix resolution $\varepsilon$
\ENSURE Closest unexplored cell $c_{u(t)}^*$\\
\STATE Initialize the dimensions of coverage matrix by Equation~\ref{eq:dims}.
\STATE Convert the robot position from the global coordinate frame into the coverage map coordinate frame.
\FOR{$c_{u(t)}^j \in \Theta_{u(t)}$}
    \STATE Find distance to all unexplored cells:
    \STATE \;\;\;\; $d^j_t = ||c_{t}^0 - c_{u(t)}^j||_2$
\ENDFOR
\STATE Select the closest unexplored cell: 
\STATE \;\;\;\; $c_{u(t)}^* = argmin_{c_{u(t)}^j \in \Theta_{u(t)}}(d^j_t)$ 
\STATE \textbf{Return} $c_{u(t)}^*$
\end{algorithmic}
\end{algorithm}

\subsection{Results and Discussion}
\subsubsection{Experiment 1: Frontier-Led Swarming versus Frontier, Greedy Swarm and Swarm Algorithms}
Tables \ref{tab:eva1}-\ref{tab:eva2} and the video demonstrations in the supplementary material present the coverage performance of our proposed algorithm. All algorithms are able to cover the whole region ($CPs$ = 100\%).The differences in turnaround time are not significant at the 95\% confidence level for the FrontierSwarm, Frontier and GreedySwarm algorithms for either the homogeneous or heterogeneous robot groups in either of the environments. However, the Swarm algorithm is significantly slower to achieve 100\% coverage than either of the other three algorithms in the larger area. This is because the random movement of this algorithm has greater redundancy of coverage in the larger environment. 

On the other hand, the group metric indicates that FrontierSwarm is maintaining a tighter formation than the Frontier approach in all experimental settings. The difference in performance is statistically significant at the 95\% confidence level. The FrontierSwarm formation is not as tight as the Swarm or GreedySwarm approaches. This is likely due to the influence of the frontier rules pulling the swarm apart a little. The order metric is greater for the Frontier approach than for FrontierSwarm, GreedySwarm and Swarm approaches. The difference is statistically significant in both cases. This makes sense because the swarming approaches should encourage ordering. All of these conclusions demonstrate that the FrontierSwarm algorithm can solve an exploration and coverage task while maintaining a small communication range between robots.

Moreover, from the FrontierSwarm group and order metric rows in the two tables, we can observe that the robots' $G$ clustering locations with respect to the average robot positions are within $R_c$ (1m and 1.49m) and the velocity differences $O$ among robots maintain low values in all cases (0.3$m.s^{-1}$ and 0.11$m.s^{-1}$). This validates the swarming quality of the presented method.

\begin{table}
	\begin{center}
		\caption{\footnotesize Experiment 1: Comparative performance of the proposed FrontierSwarm algorithm in a small area with known boundaries.}
		\label{tab:eva1}
		\begin{adjustbox}{width=0.48\textwidth}
			\small
			\begin{tabular}{|c|r|r|r|r|}
			\hline
			& \multicolumn{4}{c|}{\textit{\textbf{Homogeneous Robots}}}      \\\hline
			Metrics & \multicolumn{1}{c|}{\textit{\textbf{FrontierSwarm}}}   & \multicolumn{1}{c|}{\textit{\textbf{Frontier}}}   & \multicolumn{1}{c|}{\textit{\textbf{GreedySwarm}}}   &   \multicolumn{1}{c|}{\textit{\textbf{Swarm}}}     \\\hline
			\textit{\textbf{CP} (\%)}   &       100.00 $\pm$ 0.00        &    100.00 $\pm$ 0.00  &        100.00 $\pm$ 0.00      &  100.00 $\pm$ 0.00  
			\\\hline
			\textit{\textbf{TT} (s)}  &  130.40  $\pm$ 17.81  &     124.60 $\pm$ 5.41    &     143.40 $\pm$ 26.54   & 142.00 $\pm$ 18.02 
			\\\hline
		    \textit{\textbf{G} (m)}   &  1.00 $\pm$ 0.00 &  1.37 $\pm$ 0.01  &  0.90  $\pm$ 0.00    & 0.86 $\pm$ 0.00
			\\\hline
			\textit{\textbf{O} ($m.s^{-1}$)}  &   0.30  $\pm$ 0.05 & 0.39 $\pm$ 0.10 &  0.31 $\pm$ 0.08 &  0.29 $\pm$ 0.10 
			\\\hline
			& \multicolumn{4}{c|}{\textit{\textbf{Heterogeneous Robots}}}   \\\hline
			Metrics & \multicolumn{1}{c|}{\textit{\textbf{FrontierSwarm}}}   & \multicolumn{1}{c|}{\textit{\textbf{Frontier}}} & \multicolumn{1}{c|}{\textit{\textbf{GreedySwarm}}}   &    \multicolumn{1}{c|}{\textit{\textbf{Swarm}}}
			\\\hline
			\textit{\textbf{CP} (\%)}   &       100.00 $\pm$ 0.00        &    100.00 $\pm$ 0.00 &   100.00 $\pm$ 0.00      &     100.00 $\pm$ 0.00
			\\\hline
			\textit{\textbf{TT} (s)} &  153.60 $\pm$ 25.29 &   153.80 $\pm$ 36.21  &     153.82 $\pm$ 22.38   &   179.17 $\pm$ 23.16 
			\\\hline
		    \textit{\textbf{G} (m)}  &  1.09 $\pm$ 0.00  &  1.29 $\pm$ 0.01 &  0.86  $\pm$ 0.01   &   0.90 $\pm$ 0.00\\\hline
			\textit{\textbf{O} ($m.s^{-1}$)}  & 0.12 $\pm$ 0.03  & 0.19 $\pm$ 0.20 &  0.11 $\pm$ 0.07 &  0.12 $\pm$ 0.01
			\\\hline
		\end{tabular}
	\end{adjustbox}
	\end{center}                                 
\end{table}

\begin{table}
	\begin{center}
		\caption{\footnotesize Experiment 1: Comparative performance of the proposed FrontierSwarm algorithm in a large area with known boundaries and unknown obstacles.}
		\label{tab:eva2}
		\begin{adjustbox}{width=0.48\textwidth}
			\small
		\begin{tabular}{|c|r|r|r|r|}
			\hline
			& \multicolumn{4}{c|}{\textit{\textbf{Homogeneous Robots}}}   \\\hline
			Metrics & \multicolumn{1}{c|}{\textit{\textbf{FrontierSwarm}}}   & \multicolumn{1}{c|}{\textit{\textbf{Frontier}}}   & \multicolumn{1}{c|}{\textit{\textbf{GreedySwarm}}}    & \multicolumn{1}{c|}{\textit{\textbf{Swarm}}}     \\\hline
			\textit{\textbf{CP} (\%)}      &       100.00 $\pm$ 0.00        &    100.00 $\pm$ 0.00   &    100.00 $\pm$ 0.00     &     100.00 $\pm$ 0.00        \\\hline
			\textit{\textbf{TT} (s)}      &    233.20 $\pm$ 25.00    &  239.4 $\pm$ 41.45   &   273.70 $\pm$ 44.50    &   428.6 $\pm$ 72.04      \\\hline
			\textit{\textbf{G} (m)}  &  1.05 $\pm$ 0.01 & 1.23 $\pm$ 0.00   & 0.94 $\pm$ 0.00  & 0.88 $\pm$ 0.01  
			\\\hline
			\textit{\textbf{O} ($m.s^{-1}$)}  &   0.12  $\pm$ 0.00 & 0.16  $\pm$ 0.08  & 0.10  $\pm$ 0.00  &   0.07  $\pm$ 0.00 
			\\\hline
		
		    & \multicolumn{4}{c|}{\textit{\textbf{Heterogeneous Robots}}}   \\\hline
			Metrics & \multicolumn{1}{c|}{\textit{\textbf{FrontierSwarm}}}   & \multicolumn{1}{c|}{\textit{\textbf{Frontier}}}  & \multicolumn{1}{c|}{\textit{\textbf{GreedySwarm}}}   &    \multicolumn{1}{c|}{\textit{\textbf{Swarm}}}    \\\hline
			\textit{\textbf{CP} (\%)}      &       100.00 $\pm$ 0.00        &    100.00 $\pm$ 0.00   &       100.00 $\pm$ 0.00    &  100.00 $\pm$ 0.00      \\\hline
			\textit{\textbf{TT} (s)}      &    222.6 $\pm$ 14.30   &    213.60 $\pm$ 11.80   &   240.20 $\pm$ 38.02   &  331.40 $\pm$ 55.16   \\\hline
			\textit{\textbf{G} (m)}  &     1.07 $\pm$ 0.00  &  1.28 $\pm$ 0.00  &   0.93  $\pm$ 0.00 &  0.91  $\pm$ 0.00
			\\\hline
			\textit{\textbf{O} ($m.s^{-1}$)}  &    0.12  $\pm$ 0.02 &  0.15  $\pm$  0.01 &   0.11 $\pm$ 0.01  &  0.10 $\pm$ 0.01
			\\\hline
			
		\end{tabular}
		\end{adjustbox}
	\end{center}                                 
\end{table}

\subsubsection{Experiment 2: Frontier-led Swarming versus Multi-Robot Routing}
The next experiment compares the performance of a multi-vehicle routing (sweep) approach to our FrontierSwarm strategy. This experiment is conducted in the large environment without obstacles, as the basic sweep algorithm does not account for obstacles. The heterogeneous, seven robot setup is used. 

\begin{figure}
	\centering
	\begin{subfigure}[b]{0.48\textwidth}
		\includegraphics[width=\textwidth]{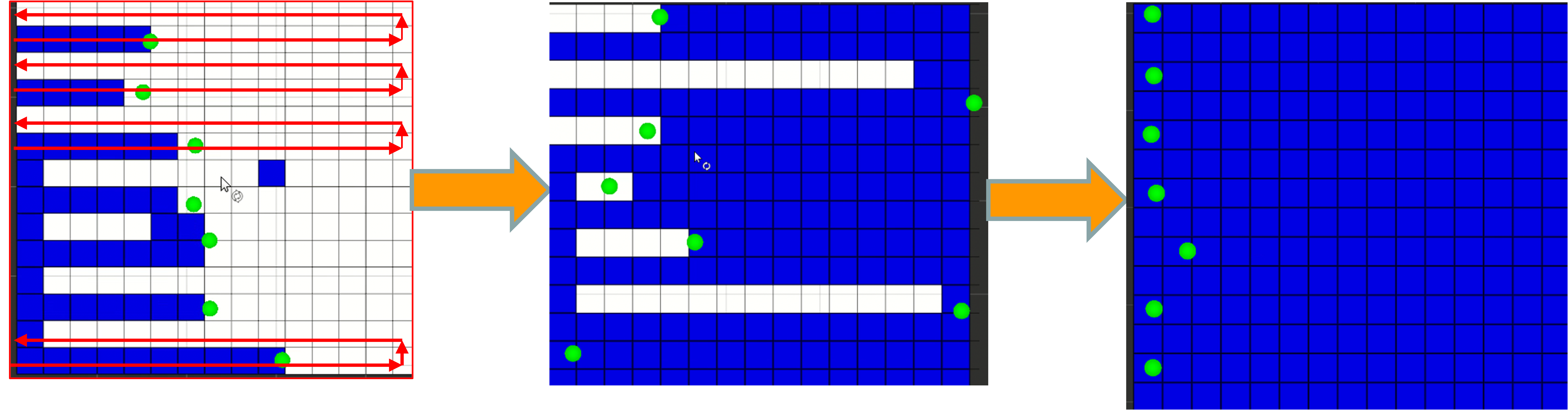}
		\caption{\textit{Multi-vehicle routing with pre-defined initial robot positions}}
	\end{subfigure}
	\begin{subfigure}[b]{0.48\textwidth}
		\includegraphics[width=\textwidth]{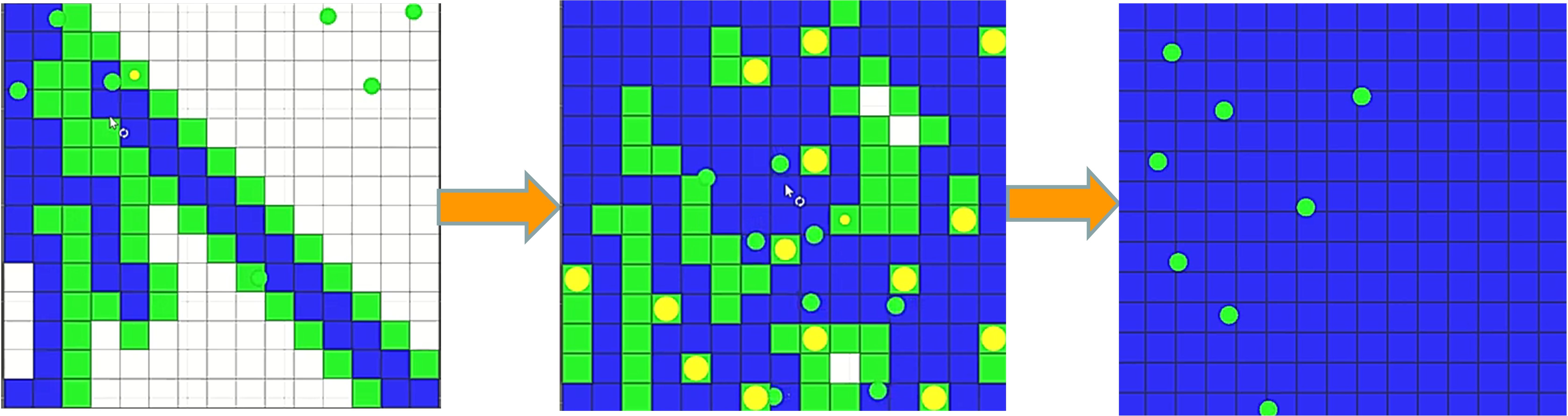}
		\caption{\textit{Frontier-led Swarming}}
		\label{fig:zn}
	\end{subfigure}
	\caption{\footnotesize Comparison of trajectories of robots during (a) sweep routing and (b) frontier-led swarming. The frontier region, the unexplored cells, the explored cells, and the chosen frontier cell are represented by green square cells, white square cells, blue square cells, and yellow circle cells, respectively.}
	\label{fig:experiment3}\vspace*{-1pt}
\end{figure}

 As can be seen in Fig. \ref{fig:experiment3}, both strategies achieve complete coverage ($CPs$ = 100\%), but the Routing approach achieves coverage in the shortest time (difference significant at the 95\% level).  This is because there is no redundancy of coverage in this approach. However, this algorithm is potentially less robust because if an individual robot fails during execution, completion of the  coverage task will be jeopardised. Moreover, it also requires prior knowledge of the structure or environment. Finally, this method can only be applied to clear regions with no obstacles. 

\begin{table}
	\begin{center}
		\caption{\footnotesize Experiment 2: Comparison of FrontierSwarm to sweep routing in an known environment without obstacles.}
		\label{tab:eva3}
		\begin{tabular}{|c|r|r|}
			\hline
			& \multicolumn{2}{c|}{\textit{\textbf{Heterogeneous Robots}}}        \\\hline
			Metrics & \multicolumn{1}{c|}{\textit{\textbf{FrontierSwarm}}}   & \multicolumn{1}{c|}{\textit{\textbf{Routing}}}       \\\hline
			\textit{\textbf{CP}} (\%) &       100.0 $\pm$ 0.0        &    100.0 $\pm$ 0.0  \\\hline
			\textit{\textbf{TT}} (s) &  227.3  $\pm$ 30.0    &   142.0 $\pm$ 5.4	\\\hline
		\end{tabular}
	\end{center}                                 
\end{table}

\subsubsection{Experiment 3: Robustness of Frontier-Led Swarming to Communication Failure or Robot Failure}
In the third experiment, during the coverage mission, two robots among a team of eight heterogeneous mobile robots are randomly either shut down or turned on after every 30s. Inter-robot communications are used by robots to validate the health status of other robots.  After an interval of 3s in which a robot has been unable to obtain a reply from another robot, the offending robot is considered `lost,' and the cohesion and alignment forces acting on the remaining robots are ignored. Separation forces continue to be applied based on LiDAR range information provided from the still active robots. This prevents the lost robot becoming a collision hazard to the other swarm members. However, if the failed robot's neighbor data is recovered, it can immediately continue the area coverage task in collaboration with its team members.

\begin{figure}
	\centering
	\includegraphics[width=19.8pc]{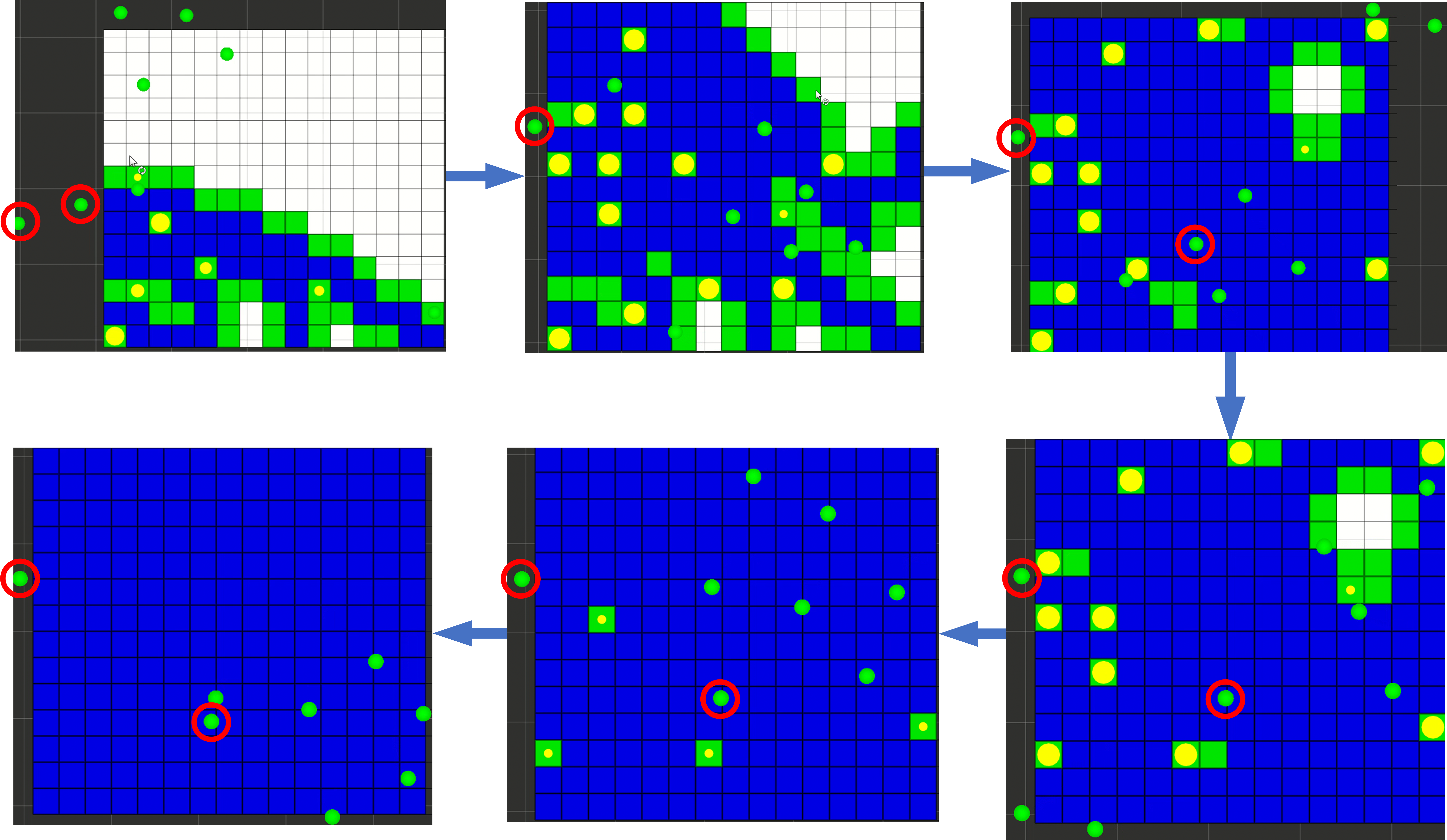}
	\caption{\footnotesize Snapshots of coverage performance over time in the presence of faulty robots (or faulty communication links). The two failed ground vehicles at any time are marked with red circles.}
	\label{fig:frontier_robust}
\end{figure}

Fig. \ref{fig:frontier_robust} shows graphically the map coverage achieved by a heterogeneous multi-robot systems using the FrontierSwarm algorithm. Six snapshots are taken during experiment. The broken robots in each snapshot are circled in red. We can see that the whole area can still be fully covered (ie. $CP$ = 100 $\%$). Moreover, although two of the robots are deactivated every 30s, the coverage completion time  $TT$ is 372 $\pm$ 24.76 s. This is significantly higher than when all robots are in full working order all the time, but the important outcome is that the coverage problem can still be solved in the presence of intermittent faults or communication failures.

\subsubsection{Experiment 4: Robustness of Frontier-Led Swarming against LiDar and Vicon Sensor Noise}

In this experiment, the robustness of the proposed coverage algorithm is examined in a cluttered environment, where noisy LiDar and VICON measurements cause the misidentification of obstacle cells, covered cells or robot position. A measurement noise level of approximately 0.1m is introduced to examine this.  

\begin{figure}
	\centering
	\begin{subfigure}[b]{0.48\textwidth}
		\includegraphics[width=\textwidth]{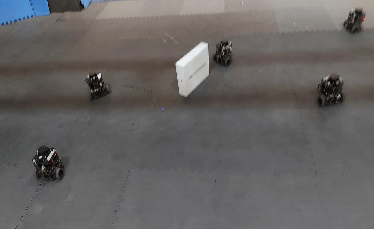}
		\caption{\textit{Real environment with one obstacle.}}
		\label{fig:env2}
	\end{subfigure}
	\begin{subfigure}[b]{0.48\textwidth}
		\includegraphics[width=\textwidth]{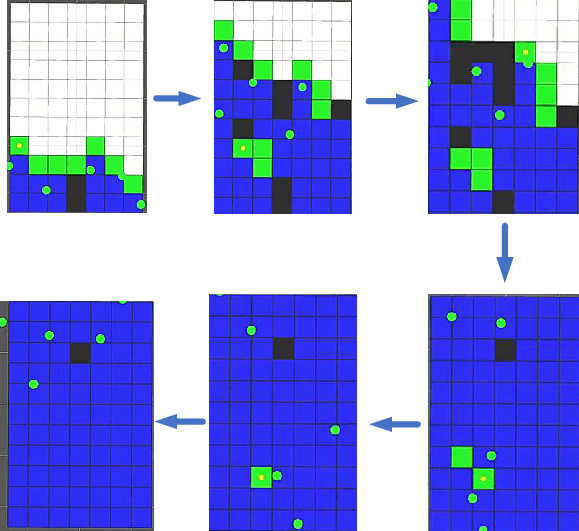}
		\caption{\textit{Estimation of object position (in black) in the presence of noisy LiDar data, visualised in the RViz interface.}}
	
	\end{subfigure}
	\caption{\footnotesize Coverage performance in the presence of noisy LiDar data.}\label{fig:exp32}
\end{figure}

As depicted in the plots of Fig. \ref{fig:exp32}(a), we establish a test environment with only one obstacle occupying a single cell. We see in Fig. \ref{fig:exp32}(b) that due to the negative effect of LiDar noise, this obstacle cell is marked at many positions early in the exploration. The incorrect obstacle cells are corrected to covered cells after more robots pass close to those cells. 

Similarly, positioning errors caused by high VICON noise levels can theoretically result in incorrect labeling of covered cells. The impact will be as follows: If the position error is less than or equal to the cell resolution size, then no incorrect record of coverage will occur. This is because the cell resolution provides some tolerance for positioning error. If the positioning error is larger than one cell resolution size, then coverage or obstacle position errors will occur in multiples of the cell resolution size.  

In this experiment with noise up to 0.1m no obstacle-robot or robot-robot collisions were recorded, and the $G$ and $O$ metrics remain within reasonable bounds of 1.05$\pm$0.078m and 0.15$\pm$0.015m, respectively. All frontier cells in Robot 1's internal map were covered, and the complete coverage task was accomplished after 220.6s.

\subsubsection{Experiment 5: Performance of the FrontierSwarm Algorithm in a Simulated, Complex Environment}

This experiment examines performance of a larger heterogeneous swarm (16 robots) in the simulated urban-like environment. Again, the experiment was repeated 5 times for each comparative algorithm. The configuration settings are identical to those used in the physical experiments (Table \ref{tab:robot}), except that the cell resolution $\varepsilon$ is increased to 1m. At the beginning of the experiment, the robot locations are evenly distributed throughout the urban map by an operator. 

As summarised in Table \ref{tab:eva4}, our proposed FrontierSwarm technique enables the robots to cover the whole unknown region with the smallest turnaround time (1522.3s for the FrontierSwarm approach compared to 1773.3s, 2234s, and 2325.7s of the Frontier, GreedySwarm and Swarm approaches, respectively). The results of the FrontierSwarm and Frontier approaches are not significantly different at the 95\% confidence level. However, the FrontierSwarm approach has perfect coverage in a similar timeframe. Furthermore, the space coverage task cannot be accomplished by the GreedySwarm approach in a reasonable coverage period. Only 96$\%$ of cells are explored within 2325.7s. It is observed that the attractive force and the repulsive force acting on each robot are often canceled out when the robot is navigating between two facing obstacles. This can cause the robot to be stationary (indecisive) for an extended period. The conclusions consolidate our arguments that the frontier-led swarming algorithm is also well-suited for larger, unknown environments. 

Despite the initial uniform distribution of robots in the large, simulated urban area, we see from Table \ref{tab:eva4} that robots using the FrontierSwarm, GreedySwarm and Swarm approaches move so that they group together and maintain close alignment during the experiments. Frontier swarm maintains closer grouping and alignment than Frontier, although has looser grouping and alignment than GreedySwarm and Swarm for the reasons discussed in Experiment 1. 

\begin{table}
	\begin{center}
		\caption{\footnotesize Experiment 5: Performance of FrontierSwarm in a cluttered urban environment with unknown obstacles.}
		\label{tab:eva4}
		 \begin{adjustbox}{width=0.48\textwidth}
		 \small
		\begin{tabular}{|c|r|r|r|r|}
			\hline
		    & \multicolumn{4}{c|}{\textit{\textbf{Heterogeneous Robots}}}   \\\hline
			Metrics & \multicolumn{1}{c|}{\textit{\textbf{FrontierSwarm}}}   & \multicolumn{1}{c|}{\textit{\textbf{Frontier}}}  & \multicolumn{1}{c|}{\textit{\textbf{GreedySwarm}}}  &    \multicolumn{1}{c|}{\textit{\textbf{Swarm}}}     \\\hline
			\textit{\textbf{CP} (\%)}      &       100.00 $\pm$ 0.00        &   98.00 $\pm$ 0.62    & 96.00  $\pm$ 5.00    &  95.00  $\pm$ 3.68   \\\hline
			\textit{\textbf{TT} (s)}      &   1522.30 $\pm$ 98.80    &  1773.30 $\pm$ 166.90   &   2234.00 $\pm$ 92.50  &   2325.70 $\pm$ 43.80    \\\hline
			\textit{\textbf{G} (m)}  &  1.16 $\pm$ 0.01  &  1.39 $\pm$ 0.01  &   1.00  $\pm$ 0.00 &  0.97  $\pm$ 0.00  \\\hline
			\textit{\textbf{O} ($m.s^{-1}$)}  &  0.34 $\pm$ 0.01 &  0.39 $\pm$ 0.00 &  0.33 $\pm$ 0.00 &  0.27 $\pm$ 0.00  
			\\\hline

		\end{tabular}
	\end{adjustbox}
	\end{center}                                 
\end{table}

\subsubsection{Experiment 6: Impact of number of robots and environment size on performance}

Table \ref{tab:deploy_time} examines the impact on turnaround time when the number of robots or size of the environment is changed. We do this systematically by first increasing the number of robots and secondly re-scaling the size of the urban-like environment. Each result is computed by averaging over ten runs. 

The first result we see is that the more robots are involved in the coverage task, the shorter the turnaround time that can be achieved. However, there appears to be evidence that the completion time may increase slightly if too many robots are used. Specifically, the 80 robot case in the 40$\times$40m takes 45s more than the 70 robots case to entirely cover the same arena. This is caused because all robots must avoid collisions and as the density of robots increases they must take more evasive actions to do this and focus less on pursuing frontier regions. In the larger environment there is evidence that increasing the number of robots will decrease the time to achieve coverage. 

\begin{table}[h]  
\caption{\footnotesize Experiment 6: Impact of number of robots and environment size on turnaround time.}\label{tab:deploy_time}
\centering 
\begin{tabular}{l c r} 
\hline\hline   
Environment Size ($m\times m$) &  No. of Robots & Turnaround Time (s)  
\\ [0.5ex]  
\hline   
$40\times 40$ & 40 & 467.25 $\pm$ 62.88  \\ 
              & 50 & 371.79 $\pm$ 80.78  \\
              & 60 & 327.60 $\pm$ 74.70 \\
              & 70 & 321.00 $\pm$ 88.77  \\
              & 80 & 366.88 $\pm$ 58.28  \\ \\[1ex] 
$80\times 80$ & 40 & 1002.42  $\pm$ 171.04  \\ 
              & 50 & 883.20 $\pm$ 123.33  \\
              & 60 & 816.81 $\pm$ 161.42 \\
              & 70 & 807.27 $\pm$ 157.07  \\
              & 80 & 725.58 $\pm$ 155.82  \\ \\[1ex] 
\hline 
\end{tabular}  
\end{table}

\section{Conclusion and Future Work}
This paper proposed a novel Frontier-led Swarming algorithm for exploration and coverage in unknown environments. Our key conclusions are:
\begin{enumerate}
   	\item Our FrontierSwarm algorithm can achieve coverage while maintaining a close-knit, self-organising formation of robots. This result was demonstrated in multiple small and larger scale environments, with different homogeneous and heterogeneous robot swarms. 
   	\item Re-tuning of algorithm parameters is not required to transfer the system to different environments.  
	\item Using a multi-ROS Master implementation and local communication, an individual robot's failure does not prevent a frontier-led swarm from completing its coverage mission, although its performance will slow. Likewise, frontier-led swarming can achieve coverage in the presence of sensor and positioning noise. 
	\item The frontier-led swarm algorithm can achieve coverage in cluttered environments with unknown obstacles, where comparative algorithms become trapped between obstacles.
\end{enumerate}

\red{A number of areas for future investigation exist, both in terms of the problem definition itself, and the solution strategy. First, the frontier-led swarming algorithm in this paper cannot track environmental changes (e.g., moving obstacle avoidance or new events) taking place in explored cells. One possible direction of future work is to combine the current approach with a digital pheromone to drive periodic revisits into explored cells to update and correct the coverage information. Secondly, this paper has not considered a search for `optimal' swarm parameters. It may be that coverage time can be decreased by trading off communication range, grouping or alignment. This could be a subject for future research using evolutionary algorithms to optimise either a single objective such as turnaround time, or multiple objectives including some of the swarming metrics studied in this paper.}

\bibliographystyle{unsrt}
\bibliography{Swarm_guidance}
\end{document}